\documentclass{article}

\usepackage[nonatbib,preprint]{neurips_2025}
\usepackage[numbers,sort,compress]{natbib}

\usepackage[utf8]{inputenc} 
\usepackage[T1]{fontenc}    
\usepackage{hyperref}       
\usepackage{url}            
\usepackage{booktabs}       
\usepackage{amsfonts}       
\usepackage{nicefrac}       
\usepackage{microtype}      
\usepackage{xcolor}         
\usepackage{graphicx}
\usepackage{algorithm}
\usepackage{algpseudocode}
\usepackage{amsmath}

\newcommand{\rj}[1]{{#1}}

\title{Task complexity shapes internal representations and robustness in neural networks}

\author{%
Robert Jankowski$^{1,2}$\thanks{Corresponding author: \texttt{robertjankowski.research@gmail.com}} \quad Filippo Radicchi$^{3}$ \quad M. Ángeles Serrano$^{1,2,4}$ \\ \textbf{Marián Boguñá}$^{1,2}$ \quad \textbf{Santo Fortunato}$^{3}$ \\ 
$^1$Universitat de Barcelona \quad $^2$Universitat de Barcelona Institute of Complex Systems (UBICS) \\ $^3$Center for Complex Networks and Systems Research (CNetS), Indiana University \quad $^4$ICREA\\
}

\begin{document}

\maketitle

\begin{abstract}
Neural networks excel across a wide range of tasks, yet remain ``black boxes''. In particular, how their internal representations are shaped by the complexity of the input data and the problems they solve remains obscure. In this work, we introduce a suite of five data-agnostic probes—pruning, binarization, noise injection, sign flipping, and bipartite network randomization—to quantify how task difficulty influences the topology and robustness of representations in multilayer perceptrons (MLPs). MLPs are represented as signed, weighted bipartite graphs from a network science perspective.
We contrast easy and hard classification tasks on the MNIST and Fashion-MNIST datasets. We show that binarizing weights in hard-task models collapses accuracy to chance, whereas easy-task models remain robust. We also find that pruning low-magnitude edges in binarized hard-task models reveals a sharp phase-transition in performance. Moreover, moderate noise injection can enhance accuracy, resembling a stochastic-resonance effect linked to optimal sign flips of small-magnitude weights. Finally, preserving only the sign structure—instead of precise weight magnitudes—through bipartite network randomizations suffices to maintain high accuracy. These phenomena define a model- and modality-agnostic measure of task complexity: the performance gap between full-precision and binarized or shuffled neural network performance. Our findings highlight the crucial role of signed bipartite topology in learned representations and suggest practical strategies for model compression and interpretability that align with task complexity.
\end{abstract}

\section{Introduction}
Neural networks have achieved remarkable success across a wide range of applications and now underpin many aspects of our daily lives. However, their vast number of trainable parameters often renders them opaque ``black boxes'' that, despite their effectiveness, sacrifice interpretability \cite{castelvecchi2016can,guidotti2018survey}. To address this, the emerging field of mechanistic interpretability (MI) seeks to reverse-engineer the parameters and algorithms of trained networks in order to understand precisely how and why they produce their outputs \cite{sharkey2025open,bereska2024mechanistic}. A common first step in MI is to decompose a network into simpler, more analyzable components. In the case of one of the simplest architectures—the multilayer perceptron (MLP)—each layer can be viewed, from a network-science perspective, as a signed, weighted bipartite graph. Such graphs are a central object of study in network science, which analyzes complex systems ranging from telecommunications and computer networks to biological and social networks~\cite{posfai2016network,newman2003structure,newman2001structure,girvan2002community}. Many real-world networks exhibit characteristic topological features—power-law degree distributions, the small-world property,  community structure, and high clustering coefficient—that reflect underlying organizational principles. By treating each MLP layer as a complex network, we can apply these same tools, such as network null models, laying the groundwork for a deeper understanding of how neural networks learn and generalize. 

Neural network performance depends not only on its architecture and training procedure but also on the complexity of the tasks it must solve~\cite{bengio2013representation,ho2022complexity}. Task difficulty shapes the representations a network learns-more challenging problems typically demand finer-grained or more abstract features~\cite{lampinen2024learned}. For example, distinguishing between visually similar classes forces the network to encode subtle differences that are not required when classes are easily separable~\cite{lin2015bilinear}. 
These differences should be observed by modeling each MLP layer as a signed bipartite graph—where positive and negative weights correspond to signed edges—and analyzing its structure.
Understanding the difficulty of a task guides model selection, architecture design, and optimization strategy. Additionally, by understanding which parts of the task are more difficult to learn, we can gain insight into how the network processes information and identify potential biases or limitations.

In this work, we investigate the internal representations learned by a fully connected multilayer perceptron (MLP) through the lens of network science, contrasting an ``easy'' task with a ``hard'' task on MNIST and Fashion-MNIST datasets. 
Conceptually, we distinguish between two related notions of difficulty. In the image experiments on MNIST and Fashion-MNIST, we use the Structural Similarity Index Measure (SSIM) solely to select representative ``easy'' and ``hard'' class pairs based on visual similarity. Our central notion of task complexity, however, is probe-based: it is defined by how a trained network’s performance degrades under controlled perturbations of its weights.
To this end, we design five complementary experimental probes: pruning (progressively removing edges with the smallest absolute weights), binarization (reducing all weights to \(\pm1\)), noise injection (adding zero-mean noise of varying amplitude to the weights), flipping signs (changing the sign of the smallest-magnitude weights), and bipartite network randomization (shuffling connections while preserving given networks' properties). 
Our key findings are as follows.
\begin{itemize}
    \item Binarizing an MLP trained on a hard task causes its accuracy to collapse to chance, whereas the easy-task model remains quite robust.
    \item As we prune low-magnitude edges, a binarized model trained on the hard task exhibits a sharp performance transition at a characteristic sparsity level.
    \item For the same model, injecting moderate noise can boost accuracy--a manifestation of stochastic resonance--while excessive noise degrades performance.
    \item The performance peak in the noise experiment arises from flipping the sign of the weights with the smallest absolute values.
    \item Randomizing the bipartite connectivity while preserving the sign of each weight leaves the network's accuracy on the easy task nearly unchanged, demonstrating that the learned representations depend more critically on the sign structure than on precise weight magnitudes.
\end{itemize}

These findings enable us to quantify task complexity in a data-agnostic manner. This means that our probes can be applied to any model and any modality as long as it contains an MLP layer.
As a case study, we used these probes to evaluate the robustness of each layer in a DistilBERT model trained for Named Entity Recognition (NER). We discovered that the earliest layers are the least robust—but that simple pruning of the smallest-magnitude weights can improve their performance. In contrast, in the deeper layers, it is the sign of each weight that matters the most. Practically speaking, this means those deeper layers can be binarized at inference time without any loss in accuracy.

\section{Related work}
\textbf{Network pruning and binary neural networks.} 
Model compression via pruning and quantization has been extensively explored to reduce inference cost while retaining accuracy~\cite{neill2020overview,Cheng2024,liang2021pruning}. Early work on Optimal Brain Surgeon (OBS) uses a second‐order Taylor approximation of the training loss to identify and remove weights with minimal impact on performance \cite{Hassibi1993Optimal}. First‐order, data‐driven methods such as Taylor pruning estimate the change in loss induced by removing individual filters or channels, achieving significant floating-point operations (FLOPs) reductions on large convolutional neural networks (CNNs) with minimal retraining \cite{Molchanov2017Pruning,He2017Channel}. More recently, single‐shot techniques like SNIP perform connection saliency scoring at initialization—eliminating the need for any gradient‐based fine‐tuning to attain high sparsity levels \cite{Lee2019SNIP}. 
Moreover, approaches based on spectral analysis, particularly those using tools from Random Matrix Theory (RMT), have been applied to characterize the spectra of weight matrices and to motivate pruning strategies~\cite{thamm2022random,staats2023boundary,staats2024small,garrod2024unifying}.
Parallel to pruning, Binary Neural Networks (BNNs) constrain weights and activations to $\{-1,+1\}$ to enable extreme compression and ultra‐fast bitwise operations. BinaryConnect and BinaryNet introduced stochastic and deterministic binarization schemes, demonstrating that end‐to‐end training of 1‐bit networks can achieve competitive accuracy on small benchmarks \cite{Courbariaux2015BinaryConnect,Rastegari2016XNOR}. Subsequent architectures such as MeliusNet employ dense feature propagation and learned scaling factors to close the gap to full‐precision models even on ImageNet-scale data \cite{Bethge2020MeliusNet}. However, these methods typically focus on worst‐case accuracy drops and rarely analyze how task difficulty modulates robustness to quantization.

\textbf{Similarity of neural network models.}
Advances in understanding how different networks or layers encode information have been driven by measures of representational and functional similarity~\cite{klabunde2023similarity}. Representational similarity measures assess how activations of intermediate layers differ, whereas functional similarity measures compare the outputs of neural networks with respect to their task. 
Representational similarity measures include Singular Vector Canonical Correlation Analysis (SVCCA)~\cite{Raghu2017SVCCA}, which aligns the subspaces spanned by activations to compare layers or models, Representational Similarity Analysis (RSA)~\cite{Kriegeskorte2008RSA}, which assesses the geometry of activation patterns by correlating pairwise distance matrices and has been applied to assess the relationship between visual tasks and their task-specific models~\cite{Kshitij2019RSA}, and Centered Kernel Alignment (CKA)~\cite{Kornblith2019CKA}, which uses kernel methods to produce robust, scale-invariant similarity scores.
On the other hand, within functional similarity measures class, we can highlight types such as: performance, hard prediction~\cite{NIPS2004_92f54963,pmlr-v119-marx20a}, soft-prediction~\cite{NIPS2014_b0c355a9}, or gradient-based~\cite{Li_2021} measures. In this work, since we work with a simple MLP, we aim to compare representation through performance analysis and probe the models' internals differently, which can be classified as one of the functional measures.

\textbf{Task complexity.}
The difficulty of learning a task has been studied from both neuroscience and machine learning perspectives. Recent empirical studies show that neural networks trained on tasks with high intra-class similarity or fine-grained distinctions tend to learn deeper or more distributed feature hierarchies~\cite{lampinen2024learned}. Additionally, Mukherjee et al.~\cite{mukherjee2020does} demonstrated that the modality of the output task plays a crucial role in shaping interpretable object representations. It has also been shown that to ensure better learning outcomes, representations may need to be tailored to both task and model to align with the implicit distribution of model and task~\cite{zilly2019quantifyingeffectrepresentationstask}.
When visually assessing images, one might intuitively conclude that datasets composed of grayscale images—such as MNIST~\cite{mnist} or Fashion-MNIST~\cite{fashionMNIST}—are generally easier to classify than RGB-valued datasets like CIFAR~\cite{cifar}. Metrics such as the Structural Similarity Index Measure (SSIM)\cite{wang2004} or Learned Perceptual Image Patch Similarity (LPIPS)\cite{zhang2018features} could serve as proxies to quantify the classification difficulty between image classes. However, in this work, we propose using neural network probes instead. These probes offer greater generalizability and can be extended beyond images to quantify task difficulty across various domains.

\textbf{Network science in deep learning}
Network science methods have been applied to neural network research in several ways. Custom loss functions based on graph‐theoretic principles have been proposed for graph neural networks \cite{bonifazi2024network}, and fully connected architectures have been analyzed in terms of classic centrality measures to link network structure with model performance \cite{scabini2023structure}. Similarly, recurrent neural networks have been shown to exhibit universal patterns of signed motifs \cite{zhang2023universal}, and more generally neural networks have been studied as dynamical systems to characterize their learning trajectories \cite{la2021characterizing, jiang2024network, lu2022understanding}.
Network science insights have also informed the design and initialization of neural architectures. Sparse connectivity patterns inspired by scale‐free graphs have been used to improve the efficiency of training large networks \cite{mocanu2018scalable}, graph‐based initialization schemes have been developed to accelerate convergence \cite{scabini2024improving}, and random wiring schemes drawn from network models have been explored~\cite{xie2019exploring}.
Graph‐theoretic metrics—such as average shortest‐path length and clustering coefficient—have been applied to characterize deep architectures, linking connectivity patterns to generalization performance \cite{you2020graph}, comparing artificial networks with biological neural circuits \cite{du2023topological}, and assessing model robustness under perturbations \cite{Waqas2022}. A recent position paper surveys many additional opportunities for network‐science approaches in deep learning \cite{blocker2025insights}.
Despite these advances, little work has applied signed, weighted bipartite graph analysis to understand how task complexity drives emergent topological transitions in the weight space. Our work fills this gap by systematically probing MLP layers under pruning, binarization, noise injection, and randomization experiments, revealing novel phase transition-like behavior dependent on task difficulty.

\textbf{Mechanistic interpretability.} 
Mechanistic interpretability has been applied primarily to large-language models, where circuit-level analyses reveal functional subnetworks and token-wise attributions~\cite{rai2024practical}. Extensions to Graph Transformers~\cite{el2025towards} and to bilinear MLPs~\cite{pearce2024bilinear} uncover attention‐based motifs and feature‐interaction circuits. However, these efforts focus on local circuits or weight factorizations, overlooking the global connectivity patterns that a network science perspective can reveal.

\section{Probing the internal representation of neural networks}
Our primary benchmarks are the MNIST and Fashion‐MNIST datasets. MNIST contains 70,000 28×28 grayscale images of handwritten digits (0–9), and Fashion‐MNIST contains 70,000 28×28 grayscale images of Zalando clothing items across 10 categories. 
In this section, the Structural Similarity Index Measure (SSIM)~\cite{wang2004} is used solely to select one easy and one hard binary classification problem for illustration; it does not enter our later definition of task complexity, which is based on the network’s response to the probes introduced below.
Initially, to define \textit{easy} and \textit{hard} tasks, we calculate the Structural Similarity Index Measure (SSIM)~\cite{wang2004} distance, i.e., $1 - \mathrm{SSIM}$, between all pairs of classes. The larger the SSIM distance, the greater the difference between the two images. Two identical images have zero SSIM distance.
We then select the class pair with the highest SSIM distance as the easy task and the pair with the lowest SSIM distance as the hard task. On MNIST, the easiest pair is \(\{0, 7\}\) and the hardest is \(\{7, 9\}\). On Fashion‐MNIST, the easiest pair is \{\texttt{Dress}, \texttt{Pullover}\}, while the hardest is \{\texttt{Dress}, \texttt{Trousers}\}. In Figure~\ref{fig:ssim} in the Appendix, we show the SSIM distance heatmaps for both datasets. Let us now define the \textbf{E-model} (\textbf{H-model}) to refer to a model trained on an easy (hard) task.

The input grayscale images are flattened, and we begin by training two multilayer perceptrons (MLPs) with ReLU activation, each with a single hidden layer of dimension $d = 64$, on binary classification tasks of differing difficulty. Optimization is performed using the Adam optimizer together with a cosine‐annealing learning‐rate schedule with a maximum of 10 epochs. At each step, we measure the test accuracy and stop training when this value is maximized. For completeness, in the Appendix, we report experiments using hidden-layer sizes $d = 32$ (Figures~\ref{fig:d32_1}-\ref{fig:d32_2}) and $d = 128$ (Figures~\ref{fig:d128_1}-\ref{fig:d128_2}) and a comparison across hidden-layer sizes in Figure~\ref{fig:prune_dim}, showing quantitatively similar trends. We also report experiments using \texttt{tanh} activation in Figure~\ref{fig:tanh}, \rj{and \texttt{hardsigmoid} in Figure~\ref{fig:hardsigmoid}, which show qualitatively similar results.}

All of our probing methods are applied to each trained network \emph{without} any further fine‐tuning or retraining. In the following sections, we provide a detailed description of each probe. The code for reproducing experiments is available at \url{https://github.com/robertjankowski/probing-neural-networks}.

\subsection{Pruning and binarizing}\label{sec:pruning}
There are many methods for pruning neural networks. In this work, we use a simple post-training strategy—iteratively removing the weights with the smallest absolute values, as in the Lottery Ticket Hypothesis~\cite{frankle2018the}. After each pruning step, we measure test accuracy on both easy and hard tasks. We also evaluate binarized versions of these pruned models—where each remaining weight is replaced by its sign—and refer to them as the signed-E and signed-H models, respectively.

\begin{figure}[h!]
    \centering
    \includegraphics[width=\textwidth]{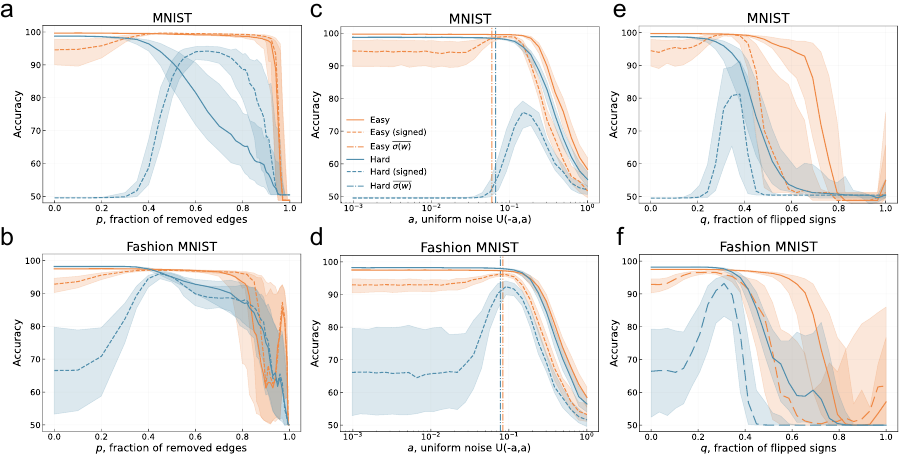}
    \caption{(\textbf{a, b}) Pruning experiment. The test accuracy as a function of the fraction of removed edges. (\textbf{c, d}) Noise injection experiment. The test accuracy as a function of the uniform noise level injected into the weights. The vertical lines show the average standard deviation of the weights. (\textbf{e, f}) Sign flipping experiment. The test accuracy as a function of the fraction of the smallest-magnitude sign flipped. All curves are averaged over 100 random initializations. Shaded regions denote the interquartile range (IQR), and the solid lines represent the median.}
    \label{fig:1}
\end{figure}

Figure~\ref{fig:1}a shows that the E-model retains higher accuracy as the smallest-magnitude weights are removed, whereas the H-model’s performance drops much more rapidly. Interestingly, pruning can even increase the accuracy of the signed-E model. Most strikingly, the signed-H model, which initially performs at near random, exhibits a performance transition and even surpasses the H-model's accuracy for some sparsity levels. A similar behavior is present for the Fashion-MNIST dataset (see Figure~\ref{fig:1}b).

Although the test accuracies of the E-model and the H-model are high, their internal representations differ. One could argue that this distinction can be measured through the distribution of weights. However, as shown in Figure~\ref{fig:std_weights}, the standard deviation of the weights, $\sigma(w)$, depends on the dataset. For MNIST, $\sigma(w)$ is narrower and smaller for the easy task, whereas it is wider and larger for Fashion-MNIST. Hence, weight statistics alone cannot serve as a reliable measure of task complexity.

In addition, in Figure~\ref{fig:strategy} in the Appendix, we compare different pruning strategies. One can observe that removing only the weights with the smallest absolute values yields a performance increase for the signed-E and signed-H models. Neither pruning the largest-magnitude weights nor randomly pruning weights exhibits this phenomenon. 
\rj{We also compare the element-wise pruning to the spectral one (see Appendix Section~\ref{sec:spectral_pruning}). Typically, only a few of the top-$k$ singular values are important for maintaining high test accuracy. Interestingly, for the signed-H-model, we observe that retaining only 2-3 singular values improves test accuracy (see Figure~\ref{fig:spectral_pruning}).}

\subsection{Noise injection}
As an additional probe, we inject noise into the network weights. Specifically, we perturb each weight $w$ by adding a random variable drawn from the uniform distribution, $U(-a,a)$, where $a$ controls the noise level.

For each noise magnitude, we evaluate test accuracy on both the E- and H-models, as well as their binarized (``signed'') counterparts. For the signed models, we first inject noise and then binarize the weights. Figures~\ref{fig:1}c,d show that the E-model remains substantially more robust under noise injection than the H-model. Moreover, adding a moderate amount of noise to the signed-E and signed-H models can actually improve their accuracy—a phenomenon akin to stochastic resonance~\cite{benzi1981,ludwig2025stochastic}, which has been documented in a wide range of systems, including bistable ring lasers, semiconductor devices, chemical reactions, and climate dynamics~\cite{Gammaitoni1998,Benzi1982,Benzi1983}.

In our context, this stochastic-resonance–like effect appears in the accuracy curves. When the noise standard deviation is much smaller than the average weight standard deviation $\bar{\sigma}(w)$ (indicated by the vertical dotted lines), we observe no improvement in model performance. Conversely, when the noise level significantly exceeds \(\bar{\sigma}(w)\), accuracy degrades to near-random levels. Thus, there exists an optimal noise-level region that maximizes performance. 

\rj{In addition, instead of additive noise, we also tested the impact of the multiplicative noise applied to a fraction of the smallest-magnitude weights (see Appendix Section~\ref{sec:multiplicative_noise}), finding that signed models remain comparatively robust and can even slightly improve for moderate perturbations when only weak-to-intermediate weights are affected (see Figure~\ref{fig:mnist_noise_multiplicative}).}

\subsection{Flipping signs}
To further investigate the stochastic resonance-like effect observed in the accuracy curves, we design a simple experiment in which we flip the signs of the smallest-magnitude weights. First, we sort all weights by their absolute values and then flip the sign of a fraction $q$ of the smallest-magnitude weights. In Figures~\ref{fig:1}e,f, we plot the test accuracy as a function of $q$ for the original models and their binarized counterparts. Consistent with our noise-injection findings, flipping a nonzero fraction of the smallest-magnitude weights in both the signed-E and signed-H models improves performance, yielding a higher-accuracy peak. These results indicate that the signs of the weights, rather than their exact values, are most critical to model performance. To test this hypothesis, we next apply a series of bipartite network randomizations.

\subsection{Bipartite network randomization}
Each MLP layer can be represented as a signed, weighted bipartite graph. The graph comprises two disjoint node sets—left $L$ (inputs) and right $R$ (outputs)—with edges only running between $L$ and $R$. A forward signal propagates from $L$ to $R$. In the unweighted case, each node $i \in L \cup R$ has two degree counts: $k_i^+$ (the number of positive‐weight edges) and $k_i^-$ (the number of negative‐weight edges). In the weighted formulation, these become strengths—$s_i^+ = \sum_{j: w_{ij}>0} w_{ij}$ and $s_i^- = \sum_{j: w_{ij}<0} |w_{ij}|$—summing the magnitudes of the positive or negative edges incident on $i$. Finally, we denote the degree (or strength) distributions over all positive and negative edges by $P(k^+)$ and $P(k^-)$ (or $P(s^+)$ and $P(s^-)$ in the weighted case). 

\begin{figure}[h!]
    \centering
    \includegraphics[width=\textwidth]{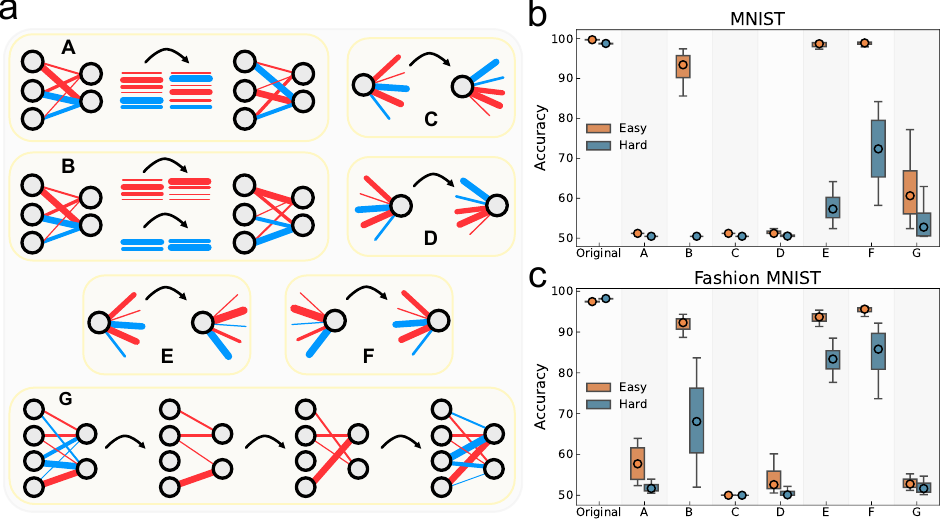}
    \caption{(\textbf{a}) Visualization of the seven types of bipartite randomizations. The accuracy of the neural network after applying each type of bipartite randomization for (\textbf{b}) MNIST and (\textbf{c}) Fashion MNIST. Boxplots show the distribution of test accuracies across 100 independent network trainings, whereas scatter markers denote the median accuracy.}
    \label{fig:2}
\end{figure}

We introduce seven distinct randomization strategies, each of which preserves different structural properties of these bipartite graphs. Figure~\ref{fig:2}a illustrates these methods, and Table~\ref{tab:1} summarizes their key characteristics. Moreover, we provide a pseudocode for each randomization in Appendix Section~\ref{apx:algorithms}.

\begin{table}[t]
    \caption{Types of bipartite randomizations and properties preserved after randomization where $\alpha$ is a fraction of the edges with a positive sign. A $\checkmark$ in the \textit{Keeps original sign} column indicates that each edge retains its original positive or negative sign under that randomization.}
    \label{tab:1}
    \centering
    \begin{small}
        \begin{tabular}{lcccccccccc}
            \toprule
            Type & $\alpha$ & $k_L^-$ & $k_L^+$ & $s_L^-$ & $s_L^+$ & $k_R^-$ & $k_R^+$ & $s_R^-$ & $s_R^+$ & Keeps original sign \\
            \midrule
            A & \checkmark & - & - & - & - & - & - & - & - & - \\
            B & \checkmark & \checkmark & \checkmark & - & - & \checkmark & \checkmark & - & - & \checkmark \\
            C & \checkmark & \checkmark & \checkmark & \checkmark & \checkmark & - & - & - & - & - \\
            D & \checkmark & - & - & - & - & \checkmark & \checkmark & \checkmark & \checkmark & - \\
            E & \checkmark & \checkmark & \checkmark & \checkmark & \checkmark & \checkmark & \checkmark & - & - & \checkmark \\
            F & \checkmark & \checkmark & \checkmark & - & - & \checkmark & \checkmark & \checkmark & \checkmark & \checkmark \\
            G & \checkmark & \checkmark & \checkmark & - & - & \checkmark & \checkmark & - & - & - \\
            \bottomrule
        \end{tabular}
    \end{small}
\end{table}

Using these randomization strategies, we evaluate the post‐randomization accuracy of our trained MLPs. Figure~\ref{fig:2}b presents accuracy boxplots for the E- and H‐models. We see that only those strategies that (1) preserve both the positive and negative degree distributions $P(k^+)$ and $P(k^-)$ and (2) retain each edge’s original sign, maintain high performance. If either of these properties is altered, accuracy falls to chance. Notably, both randomizations B and G keep the degree distributions fixed, but only B preserves accuracy—demonstrating that the specific arrangement of positive versus negative weights is itself critical.

\begin{figure}[h!]
    \centering
    \includegraphics[width=0.95\textwidth]{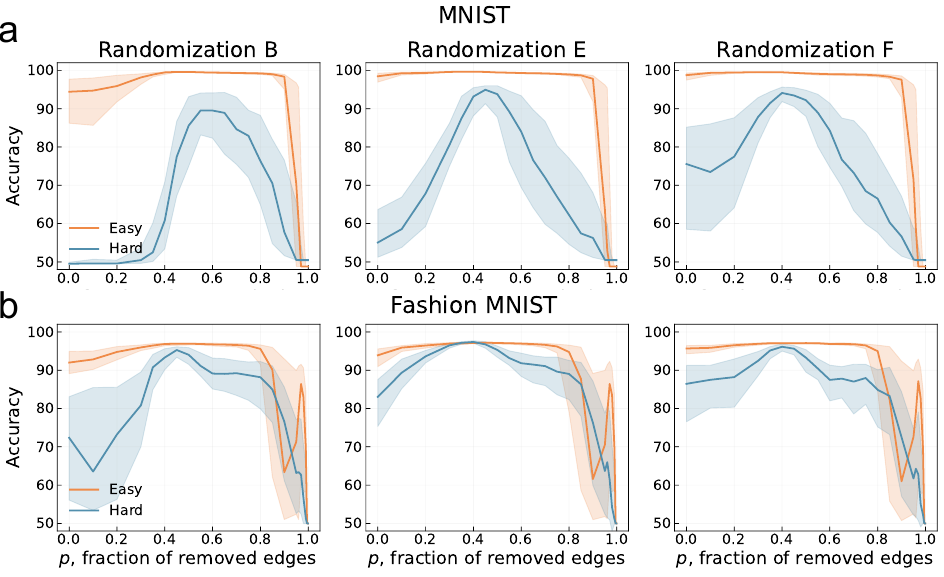}
    \caption{The test accuracy as a function of the fraction of removed edges after applying bipartite randomization B, E, and F for (\textbf{a}) MNIST and (\textbf{b}) Fashion-MNIST. All curves are averaged over 100 random initializations. Shaded regions denote the interquartile range (IQR), and the solid lines represent the median.}
    \label{fig:3}
\end{figure}

We further evaluate the randomization strategies that preserve high accuracy in the pruning experiment. As in Section \ref{sec:pruning}, we measure both the original and randomized models’ performance at each fraction of removed edges. Figure \ref{fig:3} shows that randomization initially causes an accuracy drop for both the E- and H-models. However, as sparsity increases, the randomized E-model’s accuracy steadily recovers and ultimately matches that of the original E-model. By contrast, the randomized H-model undergoes an abrupt transition—similar to the signed-H-model. These results confirm that preserving the learned edge signs, rather than the precise weight values, is essential for maintaining high performance under heavy pruning.

\section{Defining task complexity}
So far, we have compared models trained on easy and hard tasks. Binarizing or randomizing the H-model causes a large drop in accuracy, whereas the E-model’s accuracy declines much less. This suggests a link between task difficulty and post-binarization (or randomization) performance. We therefore quantify task difficulty by measuring, for each image class pair in the MNIST dataset, the change in accuracy before versus after binarizing or randomizing. 

\begin{figure}[h!]
    \centering
    \includegraphics[width=\textwidth]{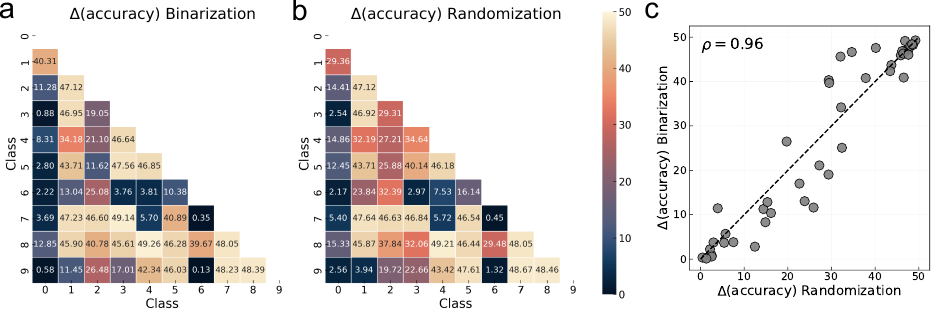}
    \caption{Difference in accuracy of the neural network for two-class discrimination under two modifications: (\textbf{a}) weight binarization, (\textbf{b}) application of bipartite randomization B. Each entry indicates the change in accuracy introduced by the modification, averaged over 10 realizations. (\textbf{c}) Scatter plot of the change in accuracy under randomization versus binarization. Each point represents a digit-class pair. In the top left corner, the Spearman correlation coefficient is reported.}
    \label{fig:4}
\end{figure}

We first note that the test accuracy for each digit class exceeds 98\% (see Figure~\ref{fig:heatmap_mnist}). Next, we evaluate how much accuracy changes once we apply our probes. In Figure~\ref{fig:4}a, we plot the difference in accuracy between each original model and its signed version. A smaller gap means the signed model’s performance remains close to the original, whereas a larger gap indicates a harder classification task. For example, digits \texttt{0} and \texttt{3} show very little change—these are easy to distinguish—while digits \texttt{1} and \texttt{7} fall to around 50\% accuracy after binarization, producing a substantial drop compared to the original. Applying bipartite randomization yields similar patterns (Figure~\ref{fig:4}b): the harder the digits are to classify, the greater the loss in accuracy. We further quantify this relation in Figure~\ref{fig:4}c. The Spearman correlation between the changes in accuracy is very high, with a coefficient of $\rho = 0.96$.

Initially, we defined \textit{easy} and \textit{hard} tasks using the Structural Similarity Index Measure (SSIM), which quantifies visual similarity between image pairs. As shown in Figure \ref{fig:4}, SSIM-easy task (\texttt{0} vs \texttt{7}) exhibits only a small drop in accuracy, whereas SSIM-hard task (\texttt{7} vs \texttt{9}) suffers accuracy losses approaching 50\%. However, SSIM requires image data. On the other hand, our approach is data-agnostic. This means that our probes can be applied to any model and any data modality, as long as it contains MLP components.

\section{Measuring layer robustness in a language model}
We emphasize that, unlike the MNIST and Fashion-MNIST experiments, we do not vary task complexity in this section; instead, we use a fixed downstream task (NER) to demonstrate that the same probes yield informative, layer-wise robustness profiles in a transformer.
In this case study, we evaluated the robustness of individual layers in a pretrained DistilBERT model\footnote{\url{https://huggingface.co/dslim/distilbert-NER}}~\cite{sanh2019distilbert}, fine-tuned on the CoNLL-2003 NER dataset~\cite{tjong-kim-sang-de-meulder-2003-introduction}. DistilBERT, a distilled variant of Bidirectional Encoder Representations from Transformers (BERT)~\cite{devlin-etal-2019-bert}, contains approximately 65 million parameters. We focus on the named entity recognition task, which aims to identify and categorize entities within text.
We apply our diagnostic probes independently to six layers of the model: (1) the positional embedding layer, (2) the first linear layer of the first transformer block, (3) the second linear layer of the first transformer block, (4) the first linear layer of the final transformer block, (5) the second linear layer of the final transformer block, and (6) the token‐classification (output) layer. For each probe, we report the test F1-score on the NER task.

\begin{figure}[h!]
    \centering
    \includegraphics[width=\textwidth]{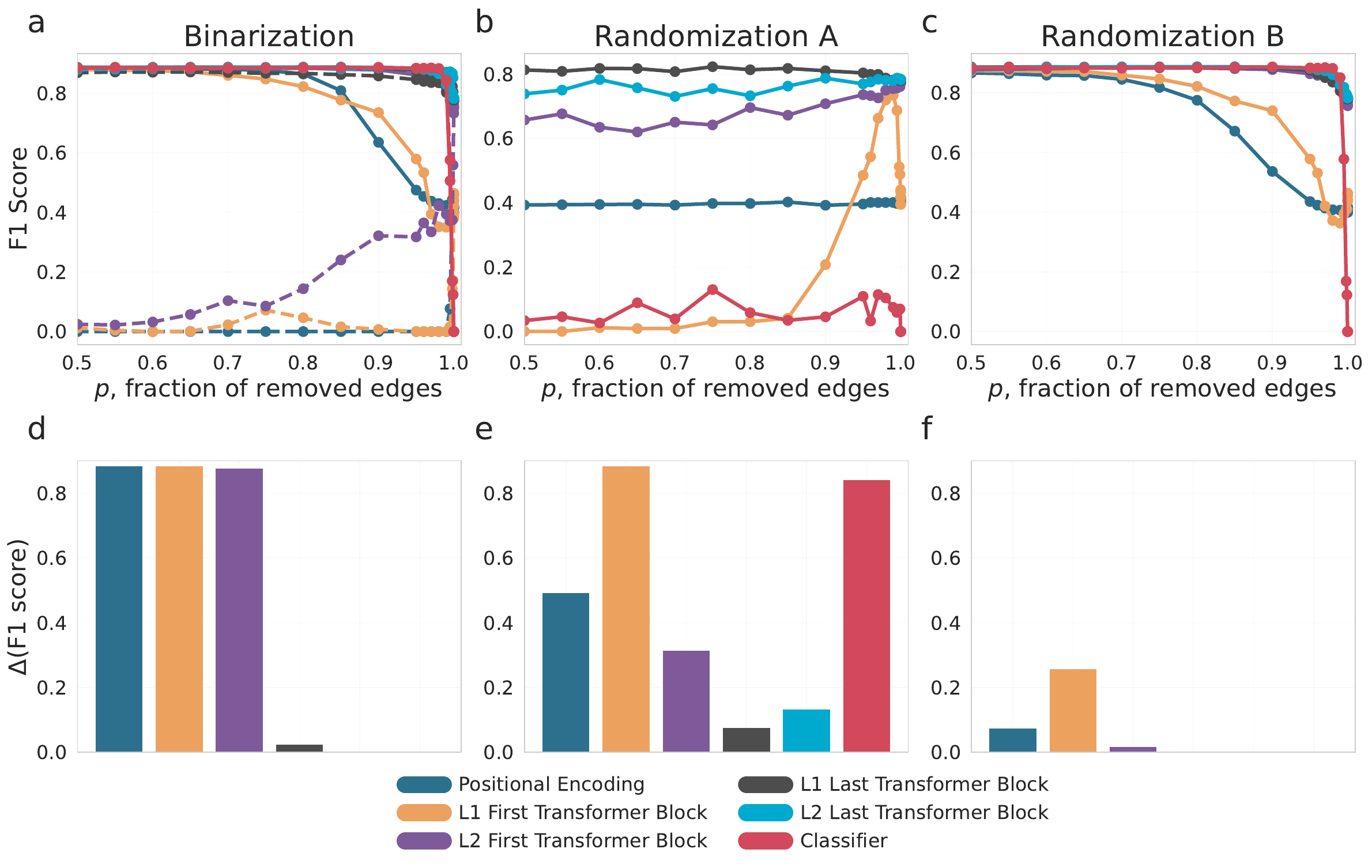}
    \caption{Case study on the DistilBERT model. F1 score as a function of the proportion of removed edges: (a) weight‐binarization experiment—solid line: original weights; dotted line: binarized weights, (b) after applying randomization~A, (c) after applying randomization~B. In the bottom panel, for $p=0$ (no edges removed), we plot the F1‐score differences under (d) weight binarization, (e) randomization~A, and (f) randomization~B. Each color corresponds to a different probed layer of DistilBERT, evaluated on the NER task.}
    \label{fig:7}
\end{figure}

In Figures~\ref{fig:7}a-c, we plot the F1 score as a function of the fraction of removed edges for six transformer layers.
First, consider the binarization experiment (Fig.~\ref{fig:7}a). The earliest layers suffer a rapid decline in F1 score as the edges are removed, with the signed model’s predictions becoming nearly indistinguishable from random. In contrast, the deepest layers remain remarkably robust: even under extreme edge removal, their binarized counterparts sustain high F1 scores. To make this comparison explicit, Figure~\ref{fig:7}d shows the difference between the original and binarized F1 curves for each layer. This pattern aligns with the findings of Bai et al.~\cite{bai-etal-2021-binarybert}, who demonstrated that shallower transformer layers are more susceptible to quantization errors than deeper ones.

Next, we examine two bipartite randomization schemes, A and B. Under randomization A (Fig.~\ref{fig:7}b), which, in isolation, previously degraded accuracy to chance, the model still retains reasonable performance when edges are removed. We attribute this resilience to the residual connections in each linear sublayer, which effectively bypass the randomized weights. Randomization B (Fig.~\ref{fig:7}c) has virtually no impact on the F1 score in the later layers, underscoring the inherent robustness of these models. 

Finally, by comparing the performance drops induced by binarization versus those induced by randomization B (Figs.~\ref{fig:7}d and~\ref{fig:7}f), we see that binarization causes a substantially larger performance drop in the early layers. 

This \rj{might be} unsurprising: perturbations at the network’s input propagate through all subsequent layers, amplifying their effect on overall performance. \rj{However, the effect could be more complex and additional tools might be needed to quantify the dumping and amplification effect, which we discuss in Appendix Section~\ref{sec:cka}.}
Randomization, by contrast, produces a more modest decline. Yet, it follows the same relative layer-wise pattern, with early layers more affected than later ones.

The same overall trends persist in the noise‐injection and sign‐flip experiments (Fig.~\ref{fig:distilbert_SI}). However, under sign flipping, the positional‐encoding layer’s performance curve becomes non‐monotonic. We believe this arises from the interplay of residual connections and LayerNorm, which together render the network invariant to a global sign inversion. Specifically, when $q=1$, we invert the entire positional-encoding vector, as the model contains six transformer blocks—an even number—each successive sign inversion is counteracted by the next, so that by the time the representation reaches the final classification head, the original encoding is effectively restored.

\section{Conclusions}
Understanding task complexity is essential for designing robust neural networks, guiding model selection, and improving training strategies. In this work, we investigated the internal representations of MLP layers by contrasting models trained on \textit{easy} and \textit{hard} tasks using five complementary probes: pruning, binarization, noise injection, sign flipping, and bipartite network randomization.

Our results show that task complexity fundamentally shapes the robustness of learned representations to perturbations. In particular, binarizing a model trained on a hard task leads to a marked loss of accuracy, whereas an easy-task model remains comparatively robust. Likewise, pruning the binarized hard-task model produces a sharp performance transition that is absent in the easy-task case. We also found that adding noise to binarized models can improve accuracy, consistent with a stochastic-resonance-like effect. 
\rj{Importantly, this performance gain from flipping a fraction of the weakest weights should not be interpreted as evidence that these individual weights are themselves the most critical to model function. Rather, the effect suggests that perturbing small weights can suppress noisy local sign assignments. More broadly, our results indicate that model performance depends less on the precise magnitudes of individual weights than on the overall organization of positive and negative connections.}
This interpretation is further supported by the bipartite network randomization experiments. In easier tasks, high accuracy is maintained only when randomization preserves both the positive/negative degree distributions and the original edge signs, confirming that the sign structure of the learned connectivity is more important than the exact floating-point weight values. These probes therefore suggest a data-agnostic way to quantify task difficulty, as the magnitude of the accuracy loss after binarization or randomization correlates with the hardness of the task or the separability of its classes.

Applying the same probes to DistilBERT on a Named Entity Recognition task revealed a different layerwise pattern: early layers are less robust to binarization, randomization, and pruning than later layers. The greater resilience of later layers, even under randomization, may be partly explained by the presence of residual connections.

Overall, our study shows that task complexity leaves a clear signature in the robustness of learned representations and highlights the central role of sign structure and connectivity in neural network function. From a practical perspective, layers whose performance is governed primarily by sign structure may be promising candidates for binarization at inference time. Future work could explore how these probe-based robustness measures relate to representational similarity metrics such as CKA~\cite{Kornblith2019CKA} or RSA~\cite{klabunde2023similarity}.

\begin{ack}
We acknowledge the support of the AccelNet-MultiNet program, a project of the National Science Foundation (Award \#1927425 and \#1927418). R.~J. acknowledges support from the fellowship FI-SDUR funded by Generalitat de Catalunya. F. R. acknowledges support from the Air Force Office of Scientific Research (Grant No.~FA9550-24-1-0039). M.~\'A.~S. and M.~B. acknowledge support from Grant No. TED2021-129791B-I00 funded by MCIN/AEI/10.13039/501100011033 and by ``European Union NextGenerationEU/PRTR'', and Grant No. PID2022-137505NB-C22 funded by MCIN/AEI/10.13039/501100011033 and by ERDF/EU.
\end{ack}

\bibliographystyle{unsrtnat}
\bibliography{ref}

\newpage
\appendix
\section{Appendix}
\subsection{Structural Similarity Index distance between pairs of classes}
\begin{figure}[h!]
    \centering
    \includegraphics[width=\textwidth]{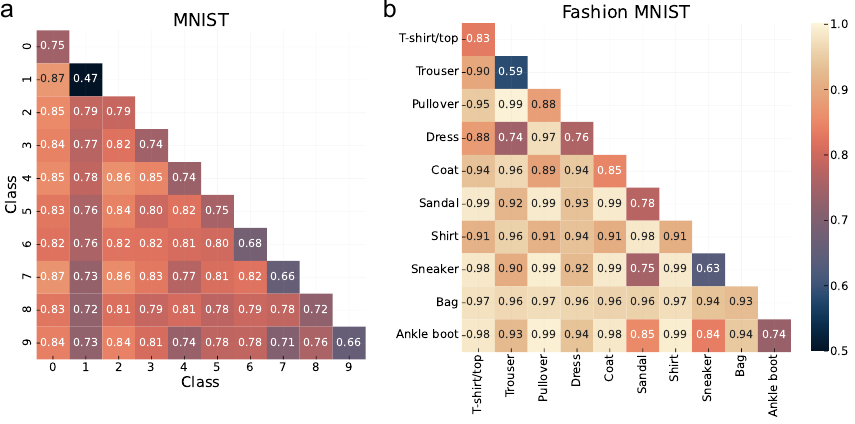}
    \caption{The Structural Similarity Index Measure (SSIM) distance between all pairs of classes for (a) MNIST and (b) Fashion MNIST. The lower the value, the more similar the two pairs of classes are.}
    \label{fig:ssim}
\end{figure}

\subsection{Standard deviations of learned weights}
\begin{figure}[h!]
    \centering
    \includegraphics[width=\textwidth]{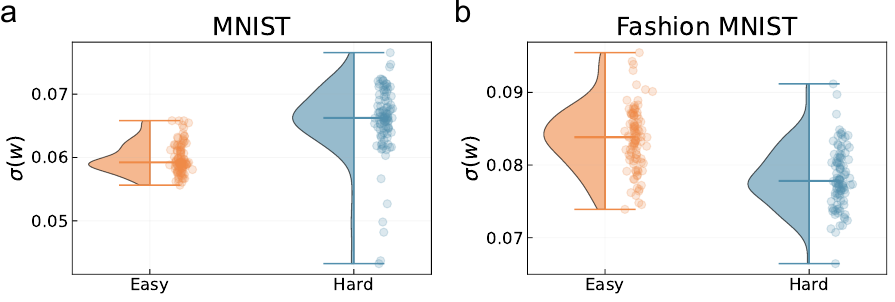}
    \caption{The distribution of weight standard deviations for (a) MNIST and (b) Fashion MNIST. Each point corresponds to a single trained neural network.}
    \label{fig:std_weights}
\end{figure}

\newpage
\subsection{Accuracy heatmap for MNIST}
\begin{figure}[h!]
    \centering
    \includegraphics[width=0.7\textwidth]{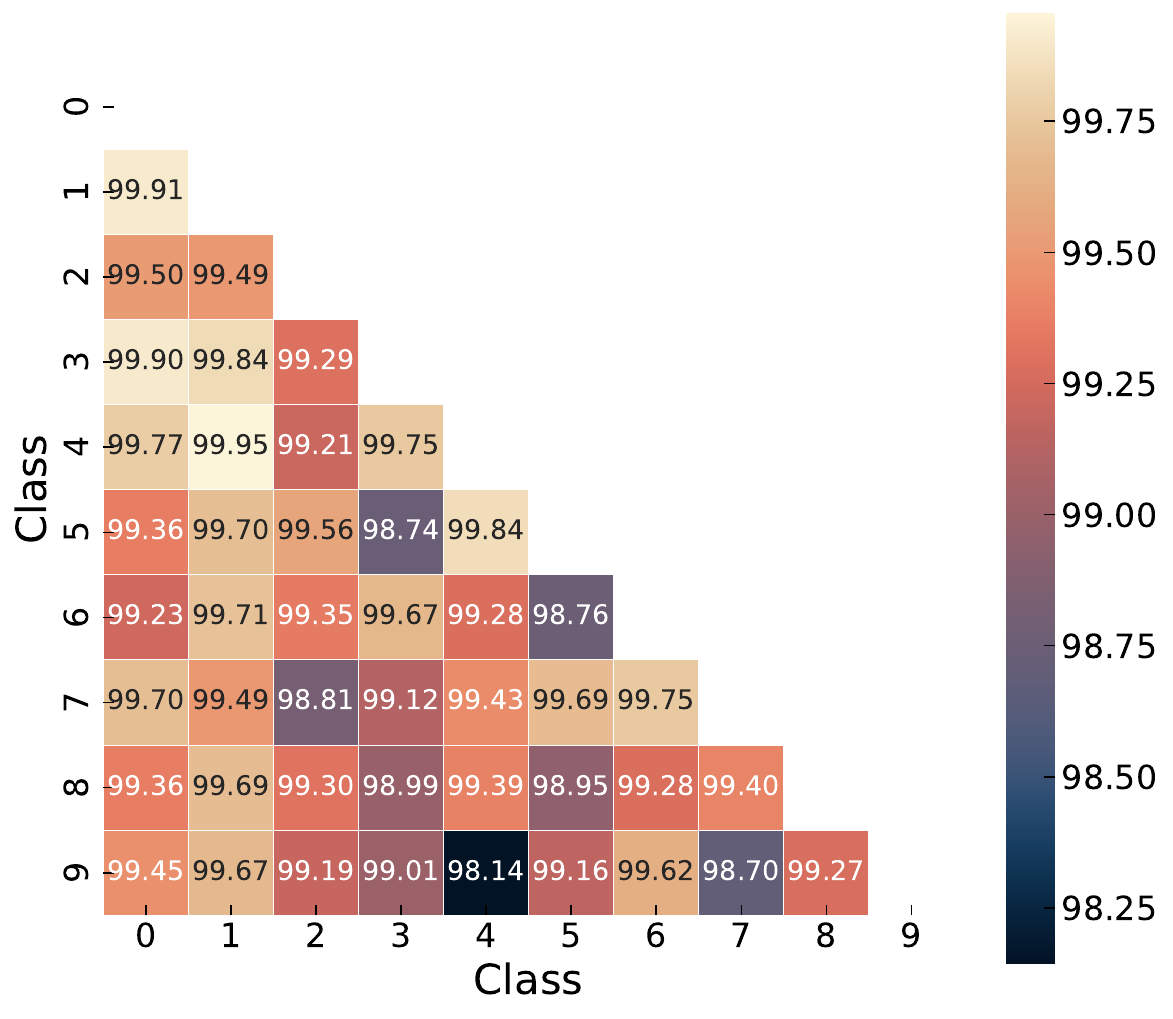}
    \caption{The accuracy heatmap for MNIST. Each entry shows the accuracy of the neural network trained to distinguish between two classes. The results are averaged over 10 realizations.}
    \label{fig:heatmap_mnist}
\end{figure}

\newpage 
\subsection{Additional experiments for DistilBERT}
\begin{figure}[h!]
    \centering
    \includegraphics[width=0.8\textwidth]{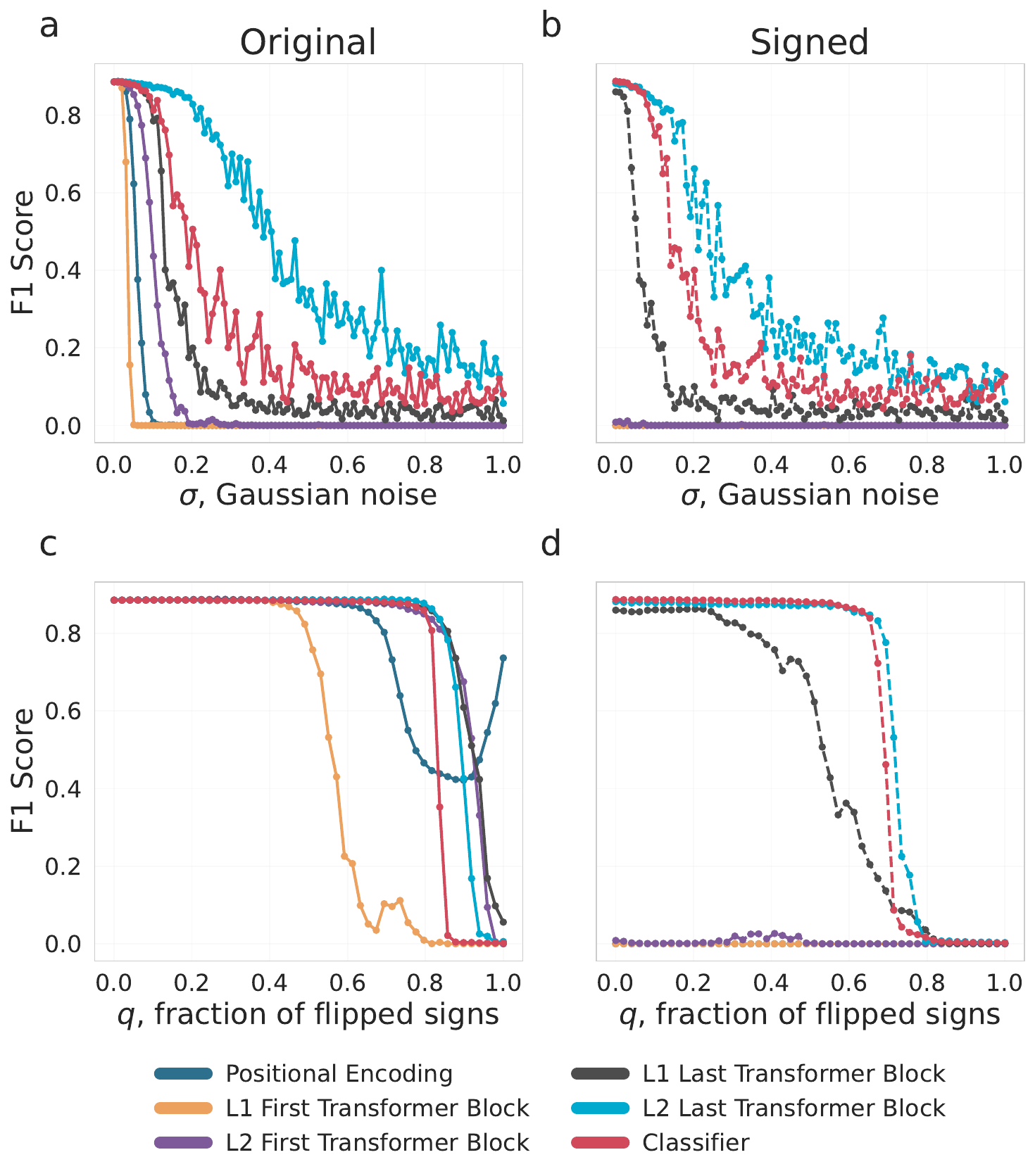}
    \caption{(a,b) The noise injection experiment. The F1 score as a function of the Gaussian noise injected into the weights. (c,d) Sign flip experiment. The F1 score as a function of the fraction of the smallest-magnitude sign flipped. Each color corresponds to a different probed layer of DistilBERT, evaluated on the NER task.}
    \label{fig:distilbert_SI}
\end{figure}

\newpage
\subsection{Additional experiments for MNIST and Fashion-MNIST datasets}

\begin{figure}[h!]
    \centering
    \includegraphics[width=\textwidth]{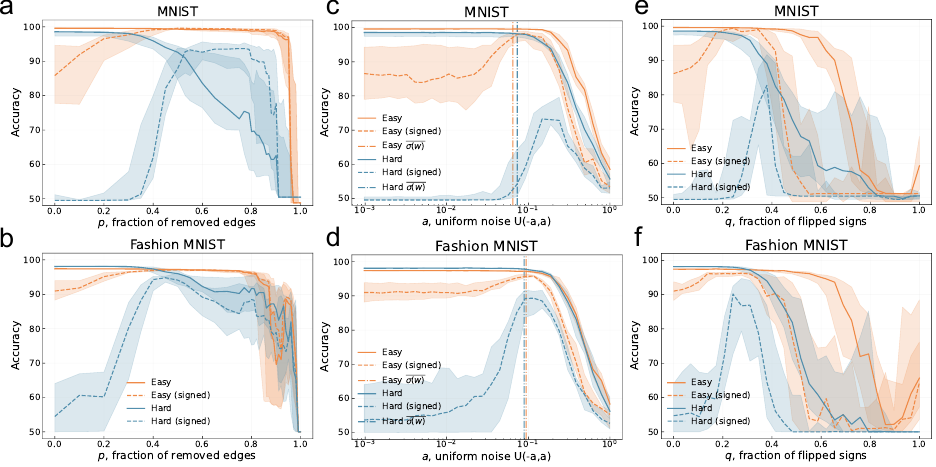}
    \caption{(\textbf{a, b}) Pruning experiment. The test accuracy as a function of the fraction of removed edges. (\textbf{c, d}) Noise injection experiment. The test accuracy as a function of the uniform noise level injected into the weights. The vertical lines show the average standard deviation of the weights. (\textbf{e, f}) Sign flipping experiment. The test accuracy as a function of the fraction of the smallest-magnitude sign flipped. All curves are averaged over 20 random initializations \textbf{for hidden layer size $d=32$}. Shaded regions denote the interquartile range (IQR), and the solid lines represent the median.}
    \label{fig:d32_1}
\end{figure}

\begin{figure}[h!]
    \centering
    \includegraphics[width=\textwidth]{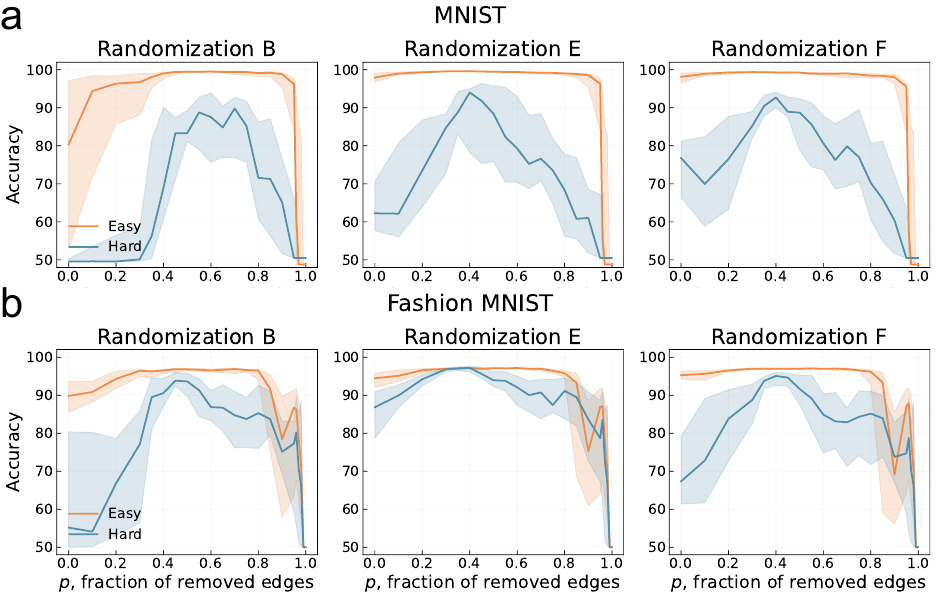}
    \caption{The test accuracy as a function of the fraction of removed edges after applying bipartite randomization B, E, and F for (\textbf{a}) MNIST and (\textbf{b}) Fashion-MNIST \textbf{for hidden layer size $d=32$}. All curves are averaged over 100 random initializations. Shaded regions denote the interquartile range (IQR), and the solid lines represent the median. }
    \label{fig:d32_2}
\end{figure}

\newpage
\begin{figure}[h!]
    \centering
    \includegraphics[width=\textwidth]{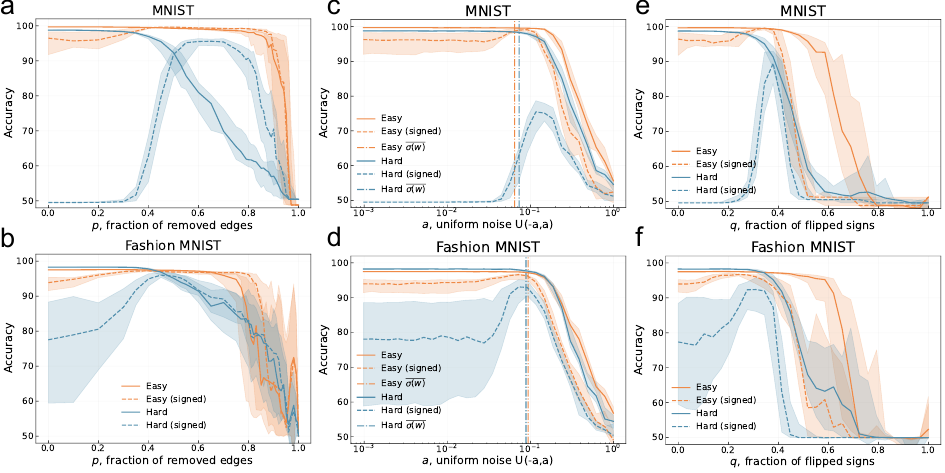}
    \caption{(\textbf{a, b}) Pruning experiment. The test accuracy as a function of the fraction of removed edges. (\textbf{c, d}) Noise injection experiment. The test accuracy as a function of the uniform noise level injected into the weights. The vertical lines show the average standard deviation of the weights. (\textbf{e, f}) Sign flipping experiment. The test accuracy as a function of the fraction of the smallest-magnitude sign flipped. All curves are averaged over 20 random initializations \textbf{for hidden layer size $d=128$}. Shaded regions denote the interquartile range (IQR), and the solid lines represent the median.}
    \label{fig:d128_1}
\end{figure}

\begin{figure}[h!]
    \centering
    \includegraphics[width=\textwidth]{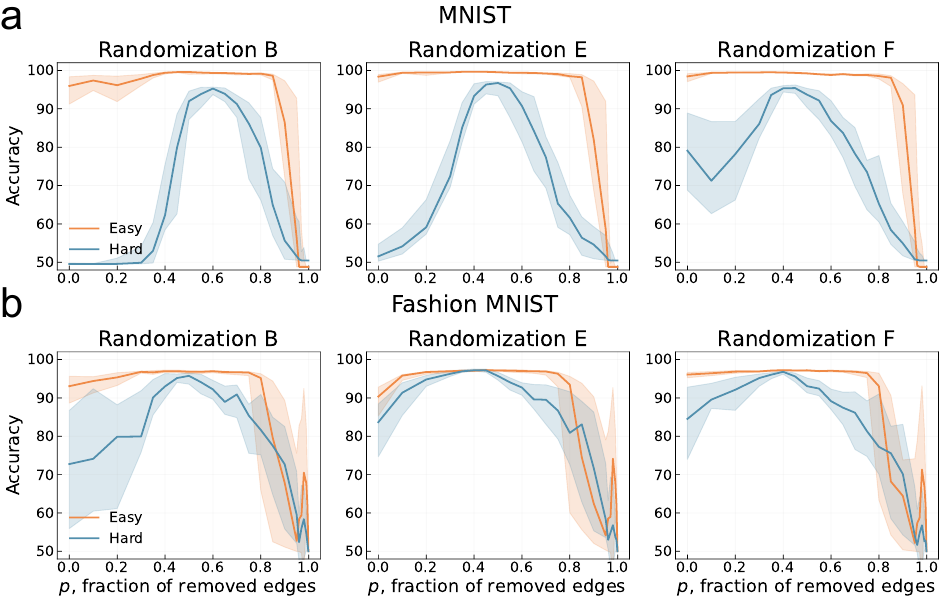}
    \caption{The test accuracy as a function of the fraction of removed edges after applying bipartite randomization B, E, and F for (\textbf{a}) MNIST and (\textbf{b}) Fashion-MNIST \textbf{for hidden layer size $d=128$}. All curves are averaged over 100 random initializations. Shaded regions denote the interquartile range (IQR), and the solid lines represent the median. }
    \label{fig:d128_2}
\end{figure}

\newpage
\begin{figure}[h!]
    \centering
    \includegraphics[width=0.85\textwidth]{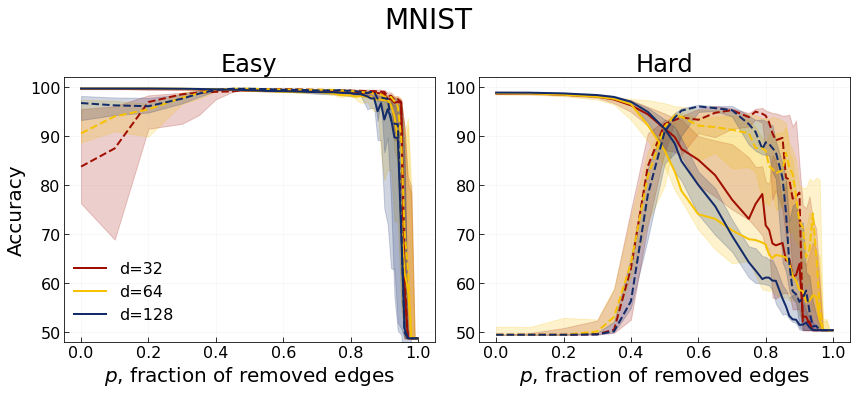}
    \caption{Comparison of different hidden dimensions in the pruning experiment for the MNIST dataset.}
    \label{fig:prune_dim}
\end{figure}

\begin{figure}[h!]
    \centering
    \includegraphics[width=\textwidth]{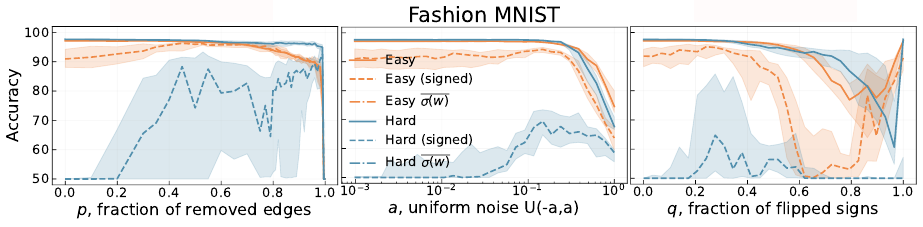}
    \caption{Experiments on Fashion-MNIST with \texttt{tanh} activation function.}
    \label{fig:tanh}
\end{figure}

\begin{figure}[h!]
    \centering
    \includegraphics[width=\textwidth]{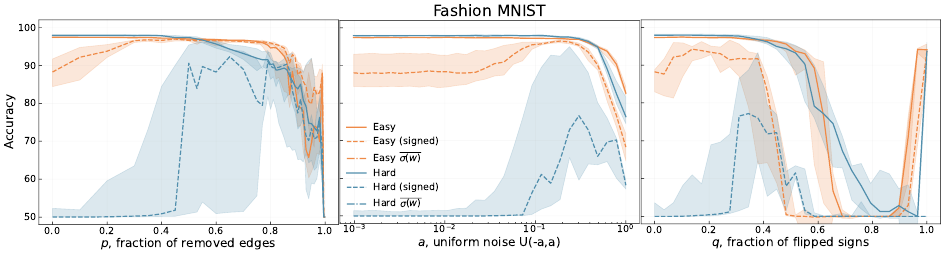}
    \caption{\rj{Experiments on Fashion-MNIST with \texttt{hardsigmoid} activation function.}}
    \label{fig:hardsigmoid}
\end{figure}

\newpage 
\begin{figure}[h!]
    \centering
    \includegraphics[width=0.9\textwidth]{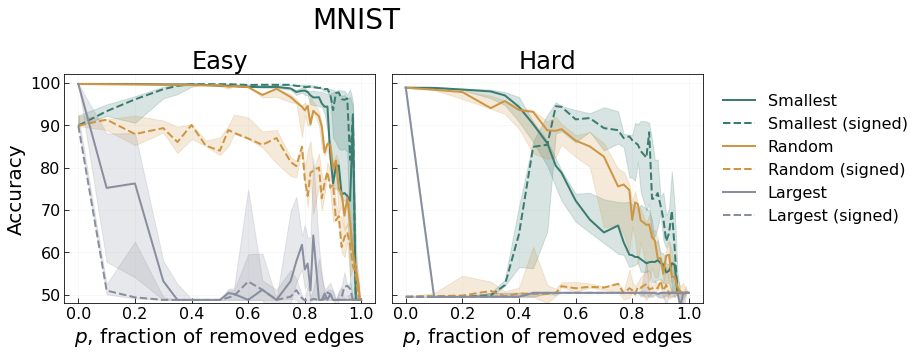}
    \caption{Pruning experiment. We apply three strategies: Smallest (removing edges with the smallest magnitude), Random (removing edges at random), and Largest (removing edges with the largest magnitude).}
    \label{fig:strategy}
\end{figure}

\subsection{Multiplicative noise}
\label{sec:multiplicative_noise}
\begin{figure}[h!]
    \centering
    \includegraphics[width=\textwidth]{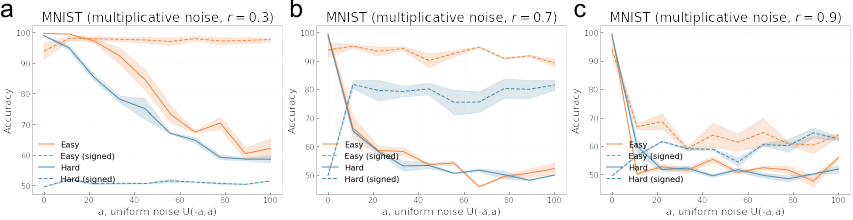}
    \caption{\rj{Multiplicative noise applied to a fraction of the smallest weights for the MNIST dataset. Panels (a)--(c) show test accuracy as a function of the noise magnitude $a$ for three values of $r$, where $r$ denotes the fraction of the smallest-magnitude weights that are perturbed.}}
    \label{fig:mnist_noise_multiplicative}
\end{figure}

\rj{Multiplicative noise is applied only to a selected fraction of the smallest-magnitude weights. Specifically, we first choose a fraction $r$ of the smallest weights, and then perturb only those weights according to}
\begin{align}
w \leftarrow w \cdot x, \qquad x \sim U(-a,a),
\end{align}
\rj{where $a$ controls the noise amplitude. We then sweep over different values of $a$ and measure the test accuracy. The results are shown in Figure~\ref{fig:mnist_noise_multiplicative} for three values of $r$.
For $r=0.3$, where only the smallest 30\% of weights are perturbed, the test accuracy of the standard models on both the Easy and Hard tasks decreases as the noise magnitude increases. By contrast, the signed model on the Easy task shows a slight improvement and remains stable even for larger values of $a$.
For $r=0.7$, we observe a similar trend for the standard Easy and Hard models. However, in this regime, the signed model on the Hard task also improves, reaching a broad plateau. This indicates that signed models can tolerate, and in some cases benefit from, perturbations affecting a substantial fraction of weak and intermediate weights.
For $r=0.9$, where all but the largest 10\% of weights are perturbed, the accuracy of all models decreases rapidly. In this case, the perturbation is no longer confined to clearly negligible weights and begins to affect increasingly important connections, which leads to a substantial degradation in performance.}

\newpage
\subsection{Spectral pruning}\label{sec:spectral_pruning}
\begin{figure}[h!]
    \centering
    \includegraphics[width=0.5\textwidth]{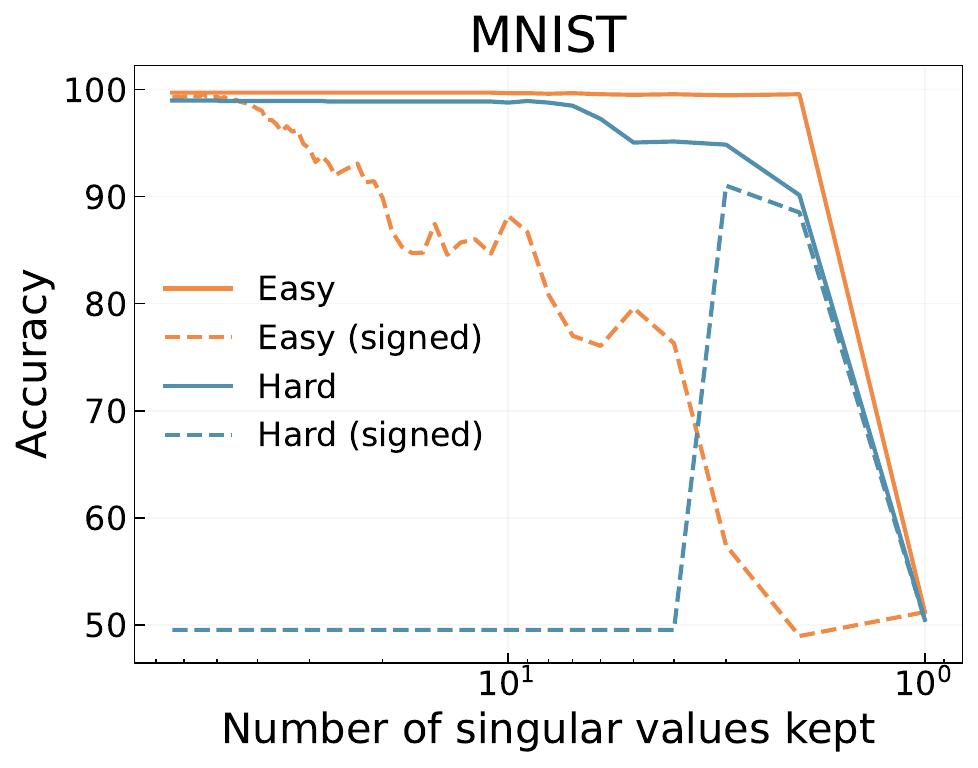}
    \caption{\rj{Spectral pruning on MNIST: test accuracy versus truncation rank $k$ (number of singular values retained). For each layer, the weight matrix is replaced by its rank-$k$ truncated SVD approximation, sweeping from the full-rank model down to $k=1$.}}
    \label{fig:spectral_pruning}
\end{figure}

\rj{For each weight matrix, we compute its singular value decomposition and construct a low-rank approximation by retaining only the top-$k$ singular values. We then sweep $k$ from the full rank down to $k=1$ and measure the test accuracy at each step.
Figure~\ref{fig:spectral_pruning} reports test accuracy as a function of the number of retained singular values for the MNIST dataset. For the Easy task, retaining only two singular values suffices to maintain high performance, whereas the Hard task requires more singular values. Interestingly, for the signed model on the Hard task, we observe a performance peak: keeping only $k=2$ or $k=3$ singular values improves test accuracy relative to higher ranks. This suggests that, in this variant, task-relevant information is concentrated in a small number of dominant singular modes, while intermediate spectral components become detrimental after binarization. A similar non-monotonic effect is observed with our element-wise pruning strategy that removes the smallest-magnitude weights (Fig.~\ref{fig:1}a,b), indicating that the improvement is not tied to a specific pruning basis but rather reflects broader redundancy in the representation. In contrast, for the Easy task, the signed model degrades gradually under spectral pruning, whereas under element-wise pruning, it shows a slight improvement.}

\subsection{CKA similarity of the transformer layers under perturbation}
\label{sec:cka}
\rj{Depending on the magnitude of the perturbation and other architectural factors, a signal can either be amplified or damped as it propagates forward. In our empirical analysis, we initially used the F1 score to quantify this effect and observed that the score deteriorates more rapidly when perturbations are applied to early layers. To study this phenomenon more directly, we propose measuring how perturbations alter the activations themselves. To this end, we use centered kernel alignment (CKA), a metric commonly used to assess similarity between neural network layers~\cite{Kornblith2019CKA}.}

\begin{figure}[h!]
    \centering
    \includegraphics[width=0.95\textwidth]{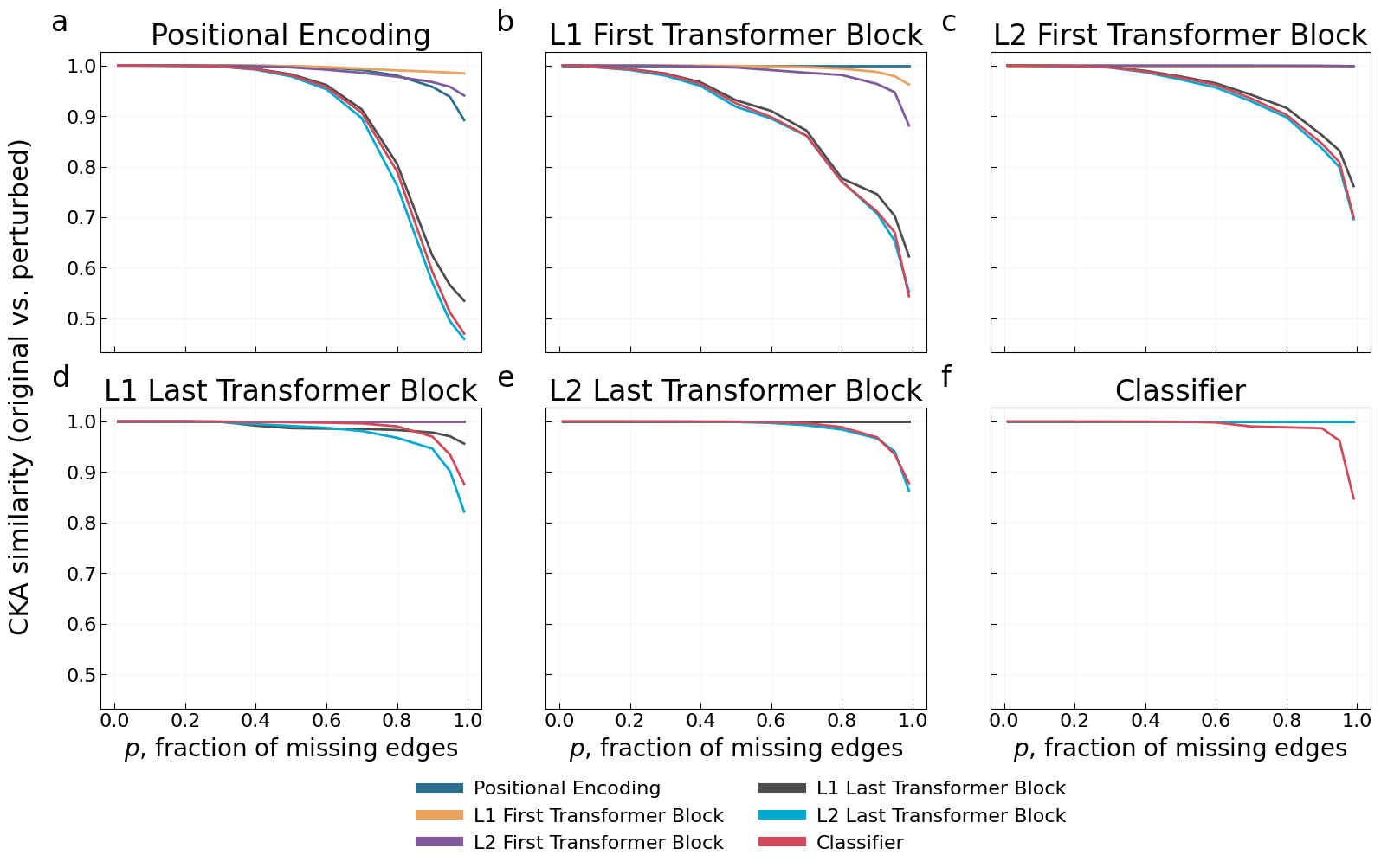}
    \caption{\rj{CKA similarity of the transformer layers under magnitude-based pruning of individual layers. Each panel (a–f) shows the linear CKA similarity between original and perturbed activations as a function of the pruning rate $p$ (fraction of edges removed) when pruning is applied only to the layer named in the panel title. Within each panel, colored curves show CKA for the activations of the layers indicated in the legend.}}
    \label{fig:cka}
\end{figure}

\rj{In Figure \ref{fig:cka}, we report the CKA similarity of the transformer layers under magnitude-based pruning of individual layers. In panel (a), for example, we perturb only the positional-encoding layer and measure how the activations of all downstream layers change. For pruning rates up to $p=0.4$, CKA remains high across all layers, indicating that the perturbation has little effect on the activations. This is consistent with our results in Figure 5a, where the F1 score also remains high for large values of $p$. At higher pruning rates, CKA similarity decreases, and different layers exhibit distinct sensitivity profiles. Interestingly, the first layer of the first transformer block maintains higher CKA similarity than the perturbed positional-encoding layer itself, suggesting that its activations are comparatively robust to perturbations in earlier components.}

\subsection{Bipartite randomization algorithms}
\label{apx:algorithms}

\begin{algorithm}[H]
\caption{Type A randomization}
\begin{algorithmic}[1]
\Require Bipartite edge list $E = \{(u,v,w_{uv})\}$
\State Separate edges into $E_0=\{w=0\}$ and $E_{\neq0}=\{w\neq0\}$
\State Extract all nonzero weights $W=\{w:(u,v,w)\in E_{\neq0}\}$
\State Randomly permute $W$
\State Reassign permuted weights back to edges in $E_{\neq0}$ (same positions)
\State \Return $E' = E_{\neq0} \cup E_0$
\end{algorithmic}
\end{algorithm}

\begin{algorithm}[H]
\caption{Type B randomization}
\begin{algorithmic}[1]
\Require Bipartite edge list $E = \{(u,v,w_{uv})\}$
\State $E^+ \gets \{(u,v,w_{uv}) \in E : w_{uv} > 0\}$  \Comment{positive edges}
\State $E^- \gets \{(u,v,w_{uv}) \in E : w_{uv} < 0\}$  \Comment{negative edges}
\State $E^0 \gets \{(u,v,w_{uv}) \in E : w_{uv} = 0\}$  \Comment{zero edges}
\State Extract weight lists $W^+,W^-$
\State Randomly permute $W^+,W^-$ separately
\State Reassign permuted weights to $E^+, E^-$ (same locations)
\State \Return $E'=E'^+ \cup E'^- \cup E^0$
\end{algorithmic}
\end{algorithm}

\begin{algorithm}[H]
\caption{Type C randomization}
\begin{algorithmic}[1]
\Require Bipartite edge list $E = \{(u,v,w_{uv})\}$
\For{each left node $u$}
    \State Split edges of $u$ into zeros and nonzeros
    \State Permute nonzero weights only within node $u$
    \State Reassign them to $u$'s edges
\EndFor
\State \Return all updated edges
\end{algorithmic}
\end{algorithm}

\begin{algorithm}[H]
\caption{Type D randomization}
\begin{algorithmic}[1]
\Require Bipartite edge list $E = \{(u,v,w_{uv})\}$
\For{each right node $v$}
    \State Separate zeros and nonzeros of node $v$
    \State Permute nonzero weights of $v$
    \State Reassign within node $v$
\EndFor
\State \Return all updated edges
\end{algorithmic}
\end{algorithm}

\begin{algorithm}[H]
\caption{Type E randomization}
\begin{algorithmic}[1]
\Require Bipartite edge list $E = \{(u,v,w_{uv})\}$
\For{each left node $u$}
    \State Collect $u$'s positive edges and negative edges separately
    \State Permute positive weights of $u$
    \State Permute negative weights of $u$
    \State Reassign them to the same positive/negative positions
\EndFor
\State \Return all edges plus zeros
\end{algorithmic}
\end{algorithm}

\begin{algorithm}[H]
\caption{Type F randomization}
\begin{algorithmic}[1]
\Require Bipartite edge list $E = \{(u,v,w_{uv})\}$
\For{each right node $v$}
    \State Collect $v$'s positive edges and negative edges separately
    \State Permute positive weights of $v$
    \State Permute negative weights of $v$
    \State Reassign them to the same positive/negative positions
\EndFor
\State \Return all edges plus zeros
\end{algorithmic}
\end{algorithm}

\begin{algorithm}[H]
\caption{Type G randomization}
\begin{algorithmic}[1]
\Require Bipartite edge list $E = \{(u,v,w_{uv})\}$
\State Extract positive edges $E^+$ and zero/negative edges $E^{\le0}$
\For{$t=1$ to number of swaps}
    \State Pick two positive edges $(u_1,v_1)$ and $(u_2,v_2)$
    \State Propose swap to $(u_1,v_2)$ and $(u_2,v_1)$
    \If{swap does not create duplicates or zero-edges}
        \State Apply the swap
    \EndIf
\EndFor
\State Reattach all negative weights in original positions (shuffled if needed)
\State Return full edge list
\end{algorithmic}
\end{algorithm}

\end{document}